\documentclass[conference,final]{IEEEtran}
\IEEEoverridecommandlockouts
\usepackage{cite}
\usepackage{amsmath,amssymb,amsfonts}
\usepackage{algorithmic}
\usepackage{graphicx}
\usepackage{textcomp}
\usepackage{xcolor}
\usepackage{tabularx}
\def\BibTeX{{\rm B\kern-.05em{\sc i\kern-.025em b}\kern-.08em
    T\kern-.1667em\lower.7ex\hbox{E}\kern-.125emX}}
\begin{document}

\title{ExClaim: Explainable Neural Claim Verification Using Rationalization}

\author{\IEEEauthorblockN{Sai Gurrapu\textsuperscript{1}, Lifu Huang\textsuperscript{1}, Feras A. Batarseh\textsuperscript{2}}
\IEEEauthorblockA{\textit{\textsuperscript{1}Department of Computer Science, \textsuperscript{2}Department of Biological Systems Engineering (BSE)} \\
\text{Virginia Tech}\\
Blacksburg, United States \\
\texttt{\{saig, lifuh, batarseh\}@vt.edu}}\\
}

\maketitle

{\let\thefootnote\relax\footnote{{Copyright and Reprint Permission: Abstracting is permitted with credit to the source. Libraries are permitted to photocopy
beyond the limit of U.S. copyright law for private use of patrons those articles in this volume that carry a code at the
bottom of the first page, provided the per-copy fee indicated in the code is paid through Copyright Clearance Center, 222 Rosewood Drive, Danvers, MA 01923. For reprint or republication permission, email to IEEE Copyrights Manager at pubs-
permissions@ieee.org. All rights reserved. Copyright  \copyright2022 by IEEE.}}}

\begin{abstract}
With the advent of deep learning, text generation language models have improved dramatically, with text at a similar level as human-written text. This can lead to rampant misinformation because content can now be created cheaply and distributed quickly. Automated claim verification methods exist to validate claims, but they lack foundational data and often use mainstream news as evidence sources that are strongly biased towards a specific agenda. Current claim verification methods use deep neural network models and complex algorithms for a high classification accuracy but it is at the expense of model explainability. The models are black-boxes and their decision-making process and the steps it took to arrive at a final prediction are obfuscated from the user. We introduce a novel claim verification approach, namely: ExClaim, that attempts to provide an explainable claim verification system with foundational evidence. Inspired by the legal system, ExClaim leverages rationalization to provide a verdict for the claim and justifies the verdict through a natural language explanation (rationale) to describe the model's decision-making process. ExClaim treats the verdict classification task as a question-answer problem and achieves a performance of 0.93 F1 score. It provides subtasks explanations to also justify the intermediate outcomes. Statistical and Explainable AI (XAI) evaluations are conducted to ensure valid and trustworthy outcomes. Ensuring claim verification systems are assured, rational, and explainable is an essential step toward improving Human-AI trust and the accessibility of black-box systems.
\end{abstract}


\section{Introduction}

\begin{figure}[ht!]
\centering
\includegraphics[scale=0.36]{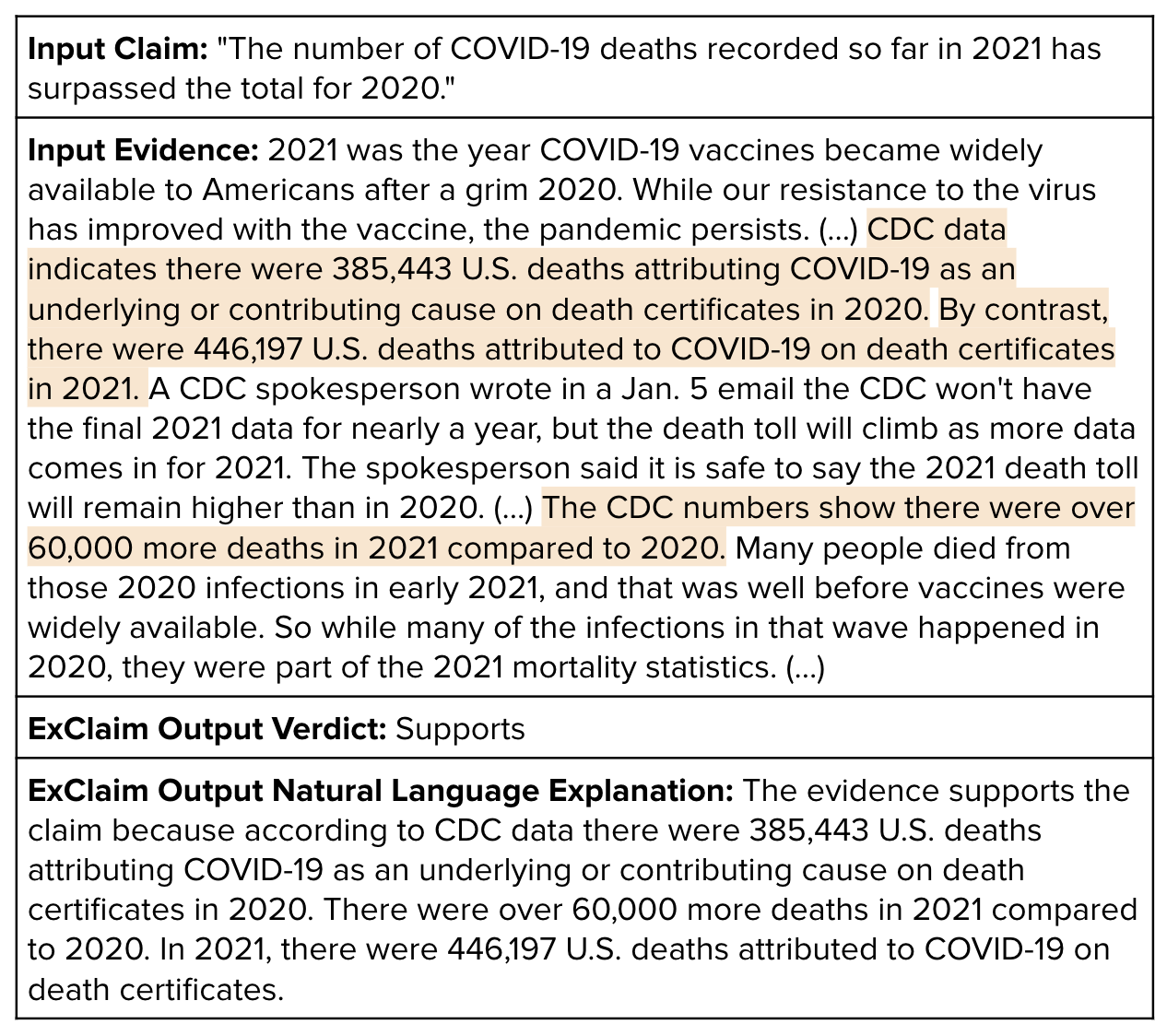}
\caption{ExClaim Approach Example}
\label{fig:exclaim_exp}
\end{figure}

Information on the web has grown remarkably in the recent few decades, and misinformation has become a significant challenge. Misinformation can weaken public trust in institutions, and economies. For instance, in 2013 Associated Press published a false tweet claiming that President Barack Obama suffered injuries from an explosion in the White House and within seconds S\&P 500 lost \$130 billion in stock value \cite{ExI1}. The impact of such information is not only on global economies but also on all aspects of our lives since it has become commonplace to consume news digitally. A leading cause for the rise of misinformation is that it can now be created cheaper and distributed faster with the internet when compared to traditional platforms such as newspapers and television \cite{ExI2}. The growth of social media usage has also allowed effortless mass sharing of information.

The recent progress with Language Models (LM) in Natural Language Processing (NLP) has made it even easier, cheaper, and faster for a machine to generate artificial text \cite{PLM}. Generative LMs have become sophisticated such as GPT-3 at text generation, and their results are almost on par with human-written text in terms of readability, coherence, and grammatical accuracy \cite{PALMLM}. In \cite{LM10x}, the authors demonstrate that LMs are improving 10x every year in terms of parameters and model size and that there is an incredible amount of progress with natural language text generation that is bound to happen. There is a possibility that these models will reach (or even surpass) human performance in a few years.

Such advances can likely lead to more rampant misinformation than what we have today online. There has been a growing interest in identifying the authenticity and truthfulness of online content. Professional fact-checkers and independent fact-checking organizations exist, but their work cannot be scaled linearly with the amount of digital content growth. This led to an increase in interest in automated fact-checking, also known as claim verification \cite{ExISurvey}. By fusing deep learning techniques with complex networks and algorithms, claim verification systems have achieved respectable performance but have become black-boxes. Optimizing for the task performance comes at the expense of model explainability. The developers of the model and the end-users which we refer to as the nonexperts do not fully know the internals or understand its decision-making processes. Furthermore, \cite{ExISurvey} points that there exists no system that offers explanations for sub-tasks in the claim verification pipeline.

Currently, a claim is fed into the system, the model checks against an evidence dataset, and it returns a binary verdict as the output; true if the evidence supports the claim otherwise false. This approach has become commonplace \cite{ExI4}; however, many questions still remain unanswered. \textit{What evidence influenced the prediction? Is the evidence trustworthy? How can the nonexpert understand the model internals?} These are all missing components of the process and there are no wide standards to assure the outcomes of these black-box systems \cite{FNSurvey}.

Explainability techniques are available but they are insufficient because many require specialized domain knowledge to understand. Claim verification systems are tools which would generally be used by a wide public audience. They need to be explainable in the sense that they appropriately justify their outcomes and are comprehensible to the nontechnical users. Local Interpretable Model-agnostic Explanations (LIME), Attention scores and saliency heatmaps are generally helpful to the model developer, but not the end-user.

An emerging NLP explainable technique is called Rationalization \cite{RXNLP}. It provides a rationale also known as a natural language explanation (NLE) to justify a model’s prediction. The reasoning behind a model’s prediction can be understood simply by reading the rationale, thereby revealing the model’s decision-making process. Rationalization can be an attractive explainable technique because it is intuitive and human-comprehensible, and does not require domain knowledge \cite{Intro4}.  It is a type of local explanation because there is a unique explanation for each prediction \cite{Intro9}. Rationalization frames explainability as an outcome-explanation problem. It looks at explainability from the perspective of an end-user whose aim is to understand how the model arrives at its final outcome \cite{Intro2}. This process is encapsulated in the rationales statement.

Rationales are also widely used in the legal system. A judge rules a verdict and it cannot be upheld without a legal opinion which is a written explanation laying out the rationale for a ruling. Inspired by this, we propose a novel approach, namely, ExClaim which leverages rationalization for an explainable neural claim verification system that is accessible and trustworthy to the nonexpert. It can not only predict a verdict but also defend and rationalize its output as a NLE. We also introduce NLP Assurance and incorporate extensive explainable evaluations to assure the verdict and NLE outcomes are valid to help reinforce Human-AI trust \cite{Assurance}. The following are the contributions of this paper:

\begin{enumerate}
    \item Introduces a new benchmark claim verification dataset with credible foundational information and a novel explainable claim verification approach called ExClaim.
    \item Uses transfer learning and presents an effective method to generate abstractive rationales in an unsupervised manner without any ground truth.
    \item Treats the verdict classification task as a question-answer problem and demonstrates that it can significantly help improve the classification performance.
    \item Presents the problem of claim verification through the lens of explainability and is the first to introduce sub-task explanations in the claim verification space.
\end{enumerate}

The structure of this paper is as follows: in Section 2, we present related work. In Section 3, we introduce a new benchmark claim verification dataset. In Section 4, we demonstrate our experiment methodology, while Section 5 describes experimental results. Lastly, in Section 6, we provide a conclusion and future directions.

\section{Related Work}
Several studies have been conducted on FEVER, LIAR, and MultiFC datasets, the most extensive datasets available for claim verification. FEVER does not contain real-world claims; instead, they are generated from Wikipedia. MultiFC uses the top \textit{k} search results from Google as its evidence for claims which is not reliable, and quality could vary drastically. Hence, we disregarded them in the study and focused on Politifact-based data. The LIAR dataset by \cite{LIAR} which was the first large-scale claim verification Politifact dataset, which led to an increase in research in this field. It is a six-way classification dataset, and \cite{LIAR} presents a Convolutional Neural Network (CNN) model that uses the claim and metadata such as speaker, party, and state, amongst others, to achieve a 0.277 accuracy. Using metadata could potentially introduce unwanted biases towards certain parties and groups. The model will be influenced by those details than using the  evidence for a claim. In \cite{ExC2}, the authors use an attention mechanism on the speaker and topic and achieve a 0.41 accuracy. 

In \cite{LIARP}, the authors publishe their LIAR-PLUS dataset with human-written justifications from each Politifact article, and their model uses a P-BiLSTM with claim and justification as inputs and achieves a 0.70 on binary classification on the Politifact data. The current SOTA performance on binary verdict classification on LIAR-PLUS is by \cite{ExC3}, and their siamese network achieves a 0.82 accuracy. Although these datasets are different from the ExClaim dataset, they shed light on the progress of verdict classification using Politifact claims. ExClaim serves as a new benchmark dataset with more credible information, and our results from Section 5 serve as the new baselines.

In the previously discussed Politifact work, the role of explainability and methods to assure the black-box predictions are untouched. A majority of explanation methods in the broader claim verification space are quantitative.  Papers \cite{ExC5}, \cite{ExC6}, \cite{ExC7}, \cite{ExC8}, and \cite{ExC9} all propose attention-based explanations using BiLSTMs, LSTMs, CNNs, co-attention networks, and decision trees, respectively. In \cite{ExC10} and \cite{ExC11}, they both use rule discovery to derive knowledge graphs and Horn rules explanations. Given the low accessibility of these explanation methods, \cite{ExC4} in 2020 presents the first study that jointly explains and predicts a verdict using a rationale. They use a supervised technique using DistilBERT for rationale extraction and perform an annotation task to evaluate rationale quality and interpretability. In \cite{ExC12}, the authors demonstrate this technique on a public health claims dataset. The progress of natural language explanations as an explainability technique has been minimal and constrained to supervised methods.

\section{ExClaim Foundational Dataset}

\begin{table}[t]
\centering
\caption{ExClaim Foundational Dataset Statistics}
\label{tab:dataset}
\begin{tabular}{lc}
\hline
\textbf{Dataset Statistics} & \textbf{}\\
\hline
Total Articles Size & 4006 \\
Training Set Articles Size & 2804 \\
Validation Set Articles Size & 601 \\ 
Test Set Articles Size & 601 \\ 
\hline
Total Support Labels & 2013 \\
Total Refute Labels & 1993  \\ 
Average Token Length of Claim & 17  \\ 
Average Token Length of Evidence & 449 \\
\hline
\end{tabular}
\end{table}

In \cite{ExISurvey}, the authors note that the availability of datasets for experimentation is the current bottleneck with automatic claim verification. The two datasets that we first considered were LIAR \cite{LIAR} and LIAR-PLUS  \cite{LIARP}. The datasets were developed by scraping Politifact, an independent fact-checking platform. However, they were not sufficient for our task because we found problems with the content with a few features missing that we required. Every Politifact article includes a claim, a six-class verdict, ruling comments written by a human fact-checker; a short justification summary at the end of the article, and metadata such as speaker, subject, and other components. In the LIAR and LIAR-PLUS datasets, they exclude the ruling comments, and previous work relies primarily on using the justification for claim verification. Frequently, this justification summary misses essential information from the verdict comments and serves as more of a closure to the article. Therefore, the datasets were not a reliable structure for our task.

We developed a new claim verification dataset using Politifact for our experiment that includes information on 4000+ articles as shown in Table \ref{tab:dataset}. Our dataset consists of the following features: Claim, Date, Source, Verdict, and the ruling comments we refer to as the Evidence and the URL. Determining the authenticity and ensuring that the evidence is neutral is crucial for claim verification. Suppose the evidence includes polarizing language and has a strong True or False sentiment. In that case, the model uses those features to predict a verdict instead of truly learning the evidence content. Additionally, Politifact often cites mainstream news mediums as sources. Most of the time, these sources are heavily biased and skew to the left or right with their content production. Due to these factors, data cleaning is performed based on the following criteria. 

When verifying claims, credible and trustworthy evidence is essential. Paper \cite{FNSurvey} points that evidence is primarily available from reputable institutions such as world governments and international organizations. News sources are mostly avenues of secondary sources. We identified the top 30 most popular news platforms and removed evidence paragraphs citing them. Our evidence includes primarily information from the U.S. government (State and Federal Laws, and government departments and agencies such as Centers for Disease Control and Prevention, Bureau of Labor Statistics, and Census Bureau amongst others) and also international organizations (United Nations,
World Health Organization, World Trade Organization and others). Politifact includes six classes for the verdict (True, Mostly True, Half True, Mostly False, False, Pants on Fire). We only scraped articles with True or False labels since the accuracy for the other classes was not clear. Claim verification systems that output \textit{True} or \textit{False} as the verdict can be misleading because there is always a possibility for error. We pre-process the verdict labels in our dataset and convert True and False into Supports and Refutes, respectively. From the end user's perspective, this approach makes it intentional that the evidence at hand either supports or refutes the claim. The end user has the option to conclude if the claim is true or false from the given outputs. We release this as the ExClaim Foundational dataset, a new benchmark claim verification dataset with foundational sources\footnote{https://github.com/AI-VTRC/ExClaim}.

\begin{figure*}[ht!]
\centering
\includegraphics[scale=0.43]{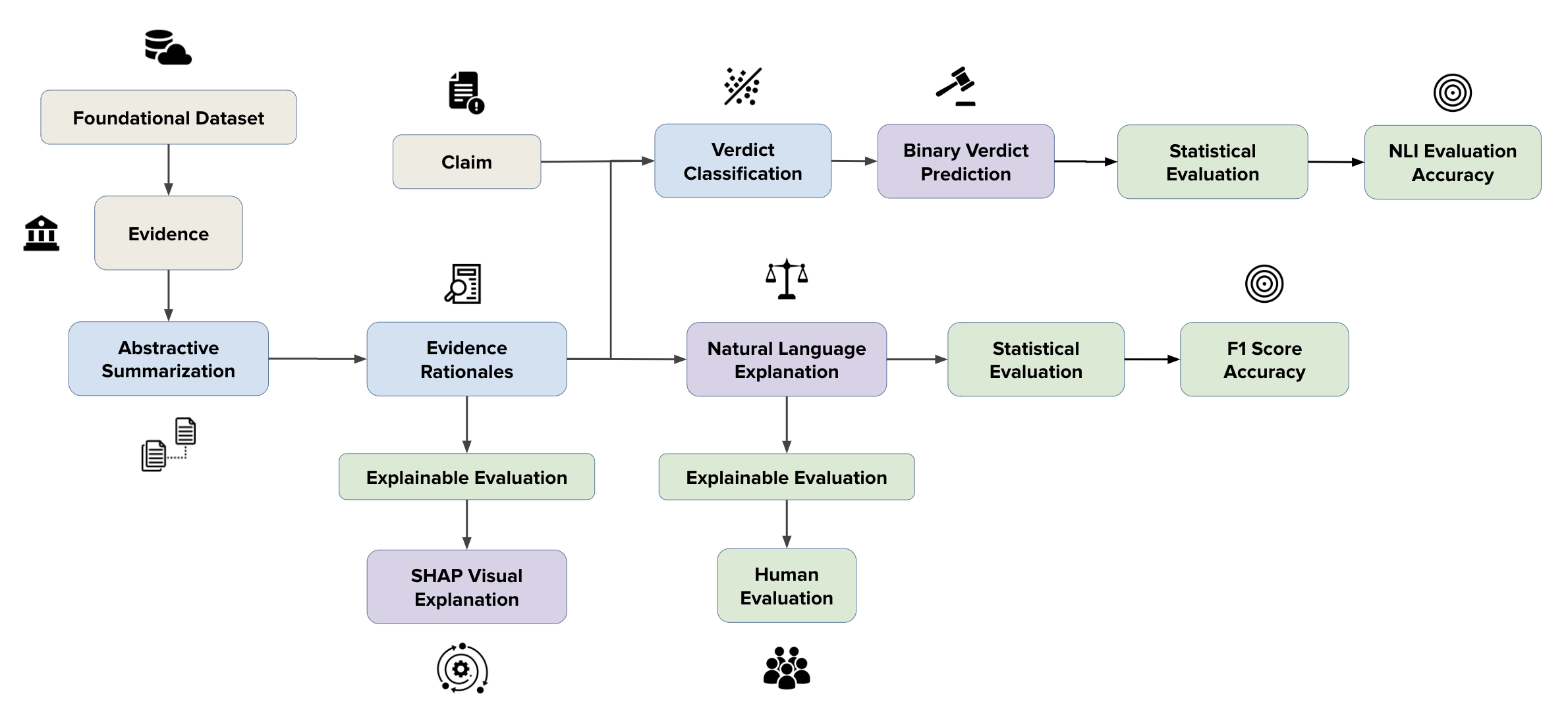}
\caption{ExClaim Architecture}
\label{fig:exclaim}
\end{figure*}

\section{Methods}

The Principle of Factor Sparsity also known as the Pareto principle states that for many outcomes in a system, 80\% of the effects come from 20\% of the causes \cite{ExM1}. This principle generally holds true in many situations and has been successfully applied in various disciplines such as medicine, engineering, and economics \cite{ExM2, ExM4}. In computer science, it was shown useful for the optimization of genetic algorithms \cite{ExM5}. It has also been applied in software testing and \cite{ExM6} demonstrates that on average "20\% of the code has 80\% of the errors”. 

Inspired by the Pareto principle, we hypothesize that on average only 20\% of the information in the evidence is important to verify a claim and the remainder 80\% is nonessential. This is similar to how humans approach manual claim verification. There are a few key pieces of information in the evidence that most influence our decision and compel us to rule that a claim is either true or false. Paper \cite{ExM13} categorizes this as an explain-then-predict model and \cite{ExM14, ExM15, ExM16} cite improvements in task performance optimization after using similar techniques. This lays the foundation for our ExClaim approach as illustrated in Figure \ref{fig:exclaim}.

\subsection{Rationale Generation}
To determine the salient information in the evidence, we use transfer learning and treat it as an unsupervised summarization task by using language models. Summarization techniques have been shown to distill the most important information from a given source with high accuracy \cite{ExM7}. We specifically use abstractive summarization because extractive techniques lack coherence, readability, and overall quality that does not reflect human-written summaries \cite{ExM8, ExM9}. We refer to the generated summary as rationales which later become part of the NLE to justify the verdict prediction.

Language models perform well at summarization as first demonstrated by \cite{Disc3} where a BERT-based abstractive summarization model outperforms most non-Transformer-based models. However, a majority of the current language models have a maximum sequence length of 512 tokens. As Table \ref{tab:dataset} shows, the average sequence length of the evidence is 449 tokens with some articles over the 512 token limit. To minimize truncation, we leverage the BART language model which has a maximum sequence length of 1024 tokens and it outperformed previous work on summarization at its publication \cite{ExM11}. BART is pre-trained on the English language and we specifically use the bart-large-cnn checkpoint from Hugging Face that has been fine-tuned on a large corpus of text-summary pairs \cite{ExM12}. We learn the function $f(E)=P^R$  where, given the evidence $E = \{E_0, E_1, …, E_n\}$, it predicts the paragraph of rationales where $P^R = \{P^R_0, P^R_1, …, P^R_n\}$. The terms $E_i$ and $P^R_i$ are the \textit{i}-th words of the evidence and the rationales paragraph, respectively.

The function $f(E)$ tokenizes the input E and passes it to a bidirectional encoder, and $P^R$ is generated through an autoregressive decoder on the final hidden layer. We use this model to perform inference on each evidence sample in the train, validation, and test sets to generate their respective rationales. The output $P^R$ is set to a maximum sequence length of 120 tokens and a minimum of 75 tokens which we found to be optimal after experimentation. The claim was not an input in this process and it is to ensure that the rationales are balanced and are not heavily biased towards any one side of the verdict.

\subsection{Verdict Classification}
In \cite{QAClass}, the authors demonstrate the benefits of multitask learning where downstream tasks are viewed as a question-answer problem using pre-trained language model.  Based on our findings, this method has not been explored in the claim verification domain. Given its promising results in other domain, we frame the ExClaim classification task as a question-answer problem. 

We leverage a pre-trained T5 model, a text-to-text generation model \cite{T5} with SOTA results on many NLP tasks including question-answering. We specifically use T5’s capability with the Choice of Plausible Alternatives (COPA) task. COPA was introduced by \cite{ExM17} and it is a type of question-answering problem that examines automated commonsense causal reasoning. The objective is given a question, a premise as context, and two answer choices perform binary classification to determine the correct answer choice. For this task, we conceive a function $g(S) = P^V$ where $S$ is the input prompt sequence and $P^V \in \{V^S, V^R\}$ is the binary verdict prediction where $V^S = \{Supports\}$ and $V^R = \{Refutes\}$. We formulate $S$ as the following where the claim as the question $C = \{C0, C1, …, Cn\}$ with $Ci$ indicating the \textit{i-th} word in the claim, the generated paragraph of rationales as the premise $P^R$, and the possible verdicts as $V^S$ and $V^R$. The sequence $S$ is fed in as “copa choice1: $V^S$ choice2: $V^R$ premise: $P^R$ question: $C$”.

We fine-tune the T5 model on our train dataset with the sequence $S$ for each sample. We set the batch size to 8, cross-entropy as the loss function and AdamW as the optimizer with a 2e-5 learning rate and we train our model for 20 epochs \cite{ExM19}. We evaluate the model’s performance on the validation set at every 350 steps. A majority of the input sequences were below T5’s maximum sequence length of 512 tokens and if the length of S went over the limit T5 does not truncate automatically since T5 uses relative positional bucketing or relative attention \cite{ExM20}. It continues to work until it is exhausted of memory which never occurred during our experimentation.

\subsubsection{Baselines}

\begin{figure*}[h]
\centering
\includegraphics[scale=0.6]{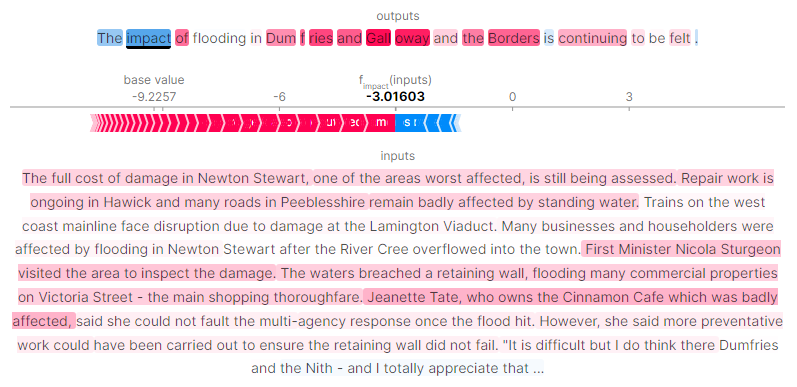}
\caption{SHAP Explanation \cite{ExR1}}
\label{fig:SHAP}
\end{figure*}

We also utilize other architectures as baselines and they are as follows.

\textbf{LSTM} Previous work such as \cite{LIAR, LIARP} have relied heavily on LSTM networks for the verdict classification. We designed a multi-input LSTM classification model. It takes the claim and evidence as two inputs which are converted into GloVe embeddings and are individually passed through an embedding layer and an LSTM layer with 128 features and the outputs of these are concatenated and passed through a dense layer with ReLu activation and a final dense layer with two neurons and softmax activation for the binary prediction output.

\textbf{DistilBERT} The DistilBERT model is also used often in the claim verification space. We fine-tune a pre-trained DistilBERT model. We set the batch size to 8, enable a sliding window to prevent truncation, use AdamW for optimization, and train for 2 epochs. 

\textbf{BART} We use a BART Zero-Shot classification model, specifically, the fine-tuned \textit{bart-large-mnli} model checkpoint from Hugging Face and we directly perform inference with a zero-shot problem setup.

\subsection{Natural Language Explanation}
The NLE is a generative task using a language model. However, during experimentation with SOTA natural language generation models such as GPT-2 \cite{GPT2} and GPT-3 \cite{GPT3}, there were challenges with respect to language model hallucination. The model hallucinates by introducing unintended and irrelevant text in the generation process. Paper \cite{ExM21} suggests that such problems diminish user trust and fail to meet their expectations. Especially for a claim verification system, introducing unintended text makes the entire system worthless and can lead to dangerous consequences. Instead, we used abstractive summarization to generate the rationales, originally it was extractive. With abstractive, the rationales reflect the natural style of a human-written text. We formulate the NLE by concatenating the claim C, the rationales $P^R$, and the verdict $P^V$. The resulting NLE, $P^N$, is as follows “The evidence $P^V$ the claim because $P^R$”. 

\section{Results and Discussion}

In this section, we describe the experimental results with supporting analysis.

\subsection{SHAP Explanation}

The process of extracting the rationales is an intermediate step in the ExClaim approach. There is also no available ground truth data to assess its accuracy. We used an explainable technique called SHAP, which leverages game theory to explain the output of any ML model \cite{ExR1}. SHAP is based on Shapley values in economics which measure the marginal contribution of a feature towards the final outcome of a system. These values are computed by meticulously perturbing the input features and observing how those changes affect the final prediction of the model. While performing inference to generate the rationales, we also pass our model to a SHAP explainer. Figure \ref{fig:SHAP} shows the output explanation where the explainer provides Shapley values for input features in the text that influenced the generation of $P^R_i$ in $P^R$. The red highlights indicate features with negative Shapley values, which are negative contributions, and similarly, blue represents positive contributions and values. SHAP is an accessible explanation method for the nonexpert because the explainer illustrates the marginal influence of each component by visualizing the Shapley values with colored highlights.

\begin{table}[h]
\centering
\caption{Macro F1 Scores for Verdict Classification Models}
\label{tab:verdictresults}
\begin{tabular}{lcc}
\hline
\textbf{Model} & \textbf{Valid} & \textbf{Test}\\
\hline
BART Zero-Shot  & - & 0.58 \\
LSTM & 0.55 & 0.47 \\
DistilBERT & 0.82 & 0.81 \\
T5 & 0.85 & 0.85 \\
ExClaim T5 & \textbf{0.92} & \textbf{0.93} \\ 
\hline
\end{tabular}
\end{table}

\subsection{Verdict Classification}

We measure the accuracy for our verdict classification models using the macro F1 score. The models we have experimented with and their results for validation and test datasets are shown in Table \ref{tab:verdictresults}. Since the BART Zero-Shot was not finetuned on this dataset, we only include the F1 score from the test set inference. Our approach using ExClaim T5 COPA outperforms the baselines, specifically, the DistilBERT model, by 10 points on the validation and 12 points on the test set. An advantage of language models is that they are pre-trained on existing knowledge. When fine-tuned on downstream tasks, they can use transfer learning and leverage the pre-existing data to improve task performance. Language models such as BART, DistilBERT, and T5 were pre-trained on vast amounts of internet data, and it hints at why the pre-trained language models performed better than the LSTM model. Amongst the language models, zero-shot classification performed poorly, and a potential reason for this is that fine-tuning generally helps improve task performance and accuracy.

\begin{table}[h]
\centering
\caption{NLI Evaluation Results}
\label{tab:nli}
\begin{tabular}{lcc}
\hline
\textbf{NLI Result} & \textbf{Count} & \textbf{Percentage}\\
\hline
Entailment  & 168 & 27.9\% \\
Neutral & 253 & 42.0\% \\
Contradiction & 180 &29.9\% \\ 
\hline
\end{tabular}
\end{table}

\subsection{Natural Language Inference }
The NLE is an open-ended text without ground truth data. This eliminated any possibility of using the ROUGE-N or BLEU score for evaluation. The work by \cite{ExR2} demonstrates that Natural Language Inference (NLI) can be confidently used to evaluate the correctness of abstractive summarization results without any human-written references. In \cite{ExR3}, they claim that assessing the accuracy of the textual entailment can be a reasonable evaluation method. NLI attempts to determine given a premise if the hypothesis is true (entailment), false (not entitlement/contradiction), or undetermined (neutral) \cite{ExR4}.

We use the NLI textual entailment technique to automatically evaluate our NLE. We used a pre-trained T5-large model, specifically, the CB task, which was fine-tuned on \cite{ExR5}’s NLI dataset. We experimented with other T5 NLI tasks such as QNLI, MNLI, and RTE, and we found that CB had the best performance. For each generated NLE, $P^N$, for the test set, we formulate the NLI input for our model as follows: “cb hypothesis: $C$ premise: $P^N$”. Essentially, we test can we logically deduce the claim from the NLE. The results are shown in Table \ref{tab:nli}. Neutral NLEs is the majority with 42\% and contradiction and entailment as the following. A potential reason for this is we are currently relying on transfer learning for the NLI evaluation, and it has not been fine-tuned on our dataset. Identifying and establishing a logical connection between the NLE and the claim can be challenging. An alternate reason is that the quality of the rationales may be sufficient enough to not be a contradiction the majority of the time but also not be sufficient enough for the premise to entail the hypothesis, which may be why neutral is the majority.

\subsection{Manual Evaluation}

In addition to the automatic evaluation with NLI, we also conduct a manual evaluation of the generated NLEs to assess its quality. For the evaluation setup, we randomly select 100 claims and its respective NLE instances from our test set. We had three annotators who were shown the claim and its NLE and their task was to judge for the following criteria: Plausibility, Fluency, and Correctness. The annotators were required to rely only on the information provided with each working individually. Since non-expert annotators might find it difficult to give thoughtful judgment, we adopt a rating scale from 1 to 5 for each criterion as presented in \ref{tab:mratings}.

We define plausibility as how convincingly can the NLE explain the verdict prediction \cite{unirex} and its rating scale is adopted from \cite{ExR7}. Fluency is defined by \cite{ExR8} as having correct spelling, good grammar, flow, and tone and its rating scale is adopted from \cite{ExR8}. Correctness is defined as the likelihood of the NLE and the verdict being true and its scale is from \cite{ExR7}.

\begin{table}[t]
\centering
\caption{Manual Evaluation Criteria and Rating Scale}
\label{tab:mratings}
\begin{tabular}{lc}
\hline
\textbf{Plausibility} & \textbf{Rating}\\
\hline
Very Convincing & 5 \\
Slightly Convincing & 4 \\
Slightly Not Convincing & 3 \\ 
Not Convincing & 2 \\ 
Can Not Judge & 1 \\ 
\hline

\textbf{Fluency} & \textbf{Rating}\\
\hline
Flawless English & 5 \\
Good English & 4 \\
Non-native English & 3 \\ 
Disfluent English & 2 \\ 
Incomprehensible & 1 \\ 
\hline

\textbf{Correctness} & \textbf{Rating}\\
\hline
Absolutely True & 5 \\
Probably True & 4 \\
Probably Not True & 3 \\ 
Absolutely Not True & 2 \\ 
Can Not Judge & 1 \\ 
\hline
\end{tabular}
\end{table}

\begin{figure}[h]
\centering
\includegraphics[scale=0.45]{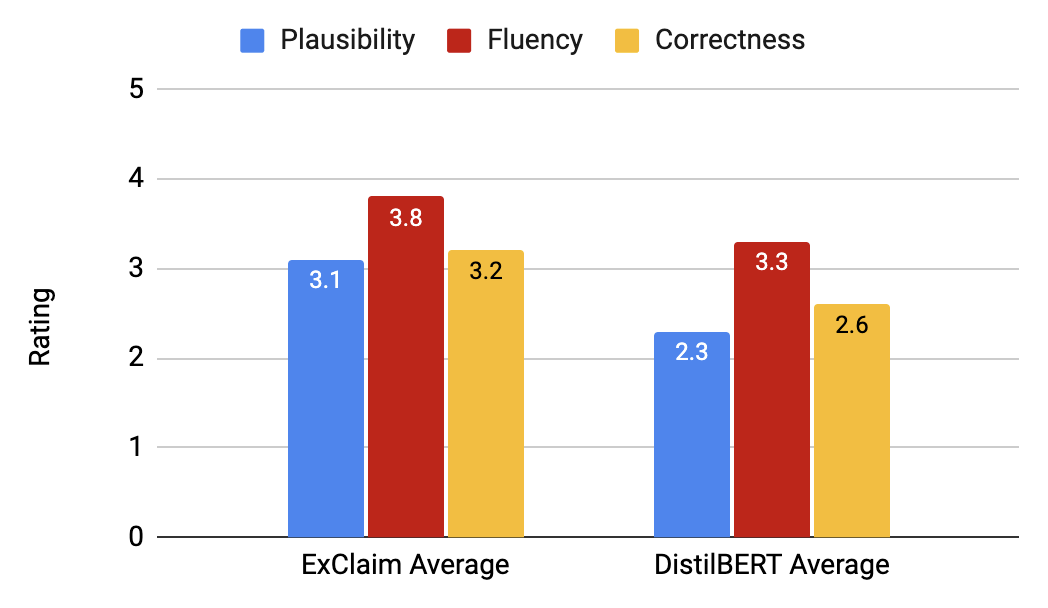}
\caption{ExClaim Manual Evaluation}
\label{fig:manev}
\end{figure}

Figure \ref{fig:manev} shows the results for the manual evaluation. We use DistilBERT as a comparison since it is used often in the claim verification space \cite{ExC4}. On average, ExClaim outperformed DistilBERT in all of the ratings. A major contributor is that ExClaim is an ensemble approach that leverages two strong language models (BART and T5) and is, therefore, better able to work in our dynamic context of rationale creation, verdict classification, and NLE generation. Both approaches have the same criteria ranking where fluency has the highest rating, then plausibility, and lastly correctness. This is expected for fluency since language models generally work well at producing proper text without major grammatical errors. Plausibility was the lowest on average and it indicates that the rationales chosen could be more persuasive. DistilBERT has a lower rating for Plausibility and it may be due to its limitations with text summarization.  The BART model used in ExClaim outperforms BERT-based models at text summarization that we used for the rationale generation. Further, DistilBERT is extractive summarization and this creates challenges with the coherency and convincingness of the rationales. Correctness on average demonstrates that there is a decent likelihood that the rationales presented with the verdict are accurate for the claim. The T5 model in ExClaim excels at transfer learning and is, therefore, better able to classify the verdict with respect to the rationales presented at inference time. In ExClaim, the verdict and rationales may have strong entailment in the NLE whereas with DistilBERT they may be more random and contradicting.

\section{Conclusions}
In this paper, we introduced a new benchmark dataset with foundational information (from government policy documents and international organizations), and we presented promising baseline results on the verdict classification task using ExClaim’s \textit{question and answer} approach. We also demonstrated an unsupervised method of using transfer learning and abstractive summarization to generate rationales which are used to determine the verdict and generate an NLE. We showed multiple explainability methods to assure the final outcomes and also the intermediate sub-tasks outcomes which has not been done before in a claim verification system. We presented a novel claim verification approach and approached the problem with rigorous explainability to ensure the system is accessible, transparent, and trustworthy to the nonexpert. 

Certain limitations do exist such as the verdict prediction and NLE being heavily impacted by the quality of the foundational data available. Additionally, claims also need to be closely related to the information included in the foundational dataset. Many promising future work directions exist to further progress in explainable claim verification. For example, developing better abstractive methods for the NLE generation from a given claim and rationales and avoiding LM hallucination. Obtaining the evidence directly from the sources listed on Politifact can yield more plausible rationales than relying on the ruling comments. Rationale generation technique without summarization has also been explored very little. Lastly, incorporating AI assurance methods as demonstrated by \cite{Assurance} in NLP is also a viable path for improving general model trustworthiness.

\bibliographystyle{IEEEtran}
\bibliography{references.bib}

\end{document}